\documentclass[sn-mathphys]{sn-jnl}
\jyear{2023}%
\begin{document}
\title[Bioinspired Soft Robotics]{Bioinspired Soft Robotics: state of the art, challenges, and future directions}

\author*[1]{\fnm{Maxwell} \sur{Hammond}}\email{maxwell-hammond@uiowa.edu}

\author[1]{\fnm{Venanzio} \sur{Cichella}}\email{venanzio-cichella@uiowa.edu}

\author[1]{\fnm{Caterina} \sur{Lamuta}}\email{caterina-lamuta@uiowa.edu}

\affil*[1]{\orgdiv{Mechanical Engineering Department}, \orgname{University of Iowa}, \orgaddress{\street{101 Jessup Hall}, \city{Iowa City}, \postcode{52240}, \state{IA}, \country{United States}}}

\abstract{

\textbf{Purpose of Review:} This review provides an overview of the state of the art in bioinspired soft robotics with by examining advancements in actuation, functionality, modeling, and control. 
\\ \textbf{Recent Findings:} Recent research into actuation methods, such as artificial muscles, have expanded the functionality and potential use of bioinspired soft robots. Additionally, the application of finite dimensional models has improved computational efficiency for modeling soft continuum systems, and garnered interest as a basis for controller formulation.
\\ \textbf{Summary:} Bioinspiration in the field of soft robotics has led to diverse approaches to problems in a range of task spaces. In particular, new capabilities in system simplification, miniaturization, and untethering have each contributed to the field's growth. There is still significant room for improvement in the streamlining of design and manufacturing for these systems, as well as in their control.
}

\maketitle

\section{Introduction}
 Evolution has advanced biological systems far beyond many capabilities of modern technology. This has created an abundant source of inspiration for the development of robotics, and recent innovations within the space of bioinspired systems have led to the establishment of soft robotics as a field with strong research interest. While the foundations of this field have existed for quite some time, with examples such as the McKibben braided pneumatic artificial muscle reaching back to the 1960s \cite{nickel1963development}, consistent use of soft robotic terminology in research has only come about in the last two decades \cite{bao2018bibliometric,laschi2016science}.  As a result of the field's youth and rapid growth, there is some debate concerning the definition of a soft robot. Early definitions of the term, which include robots made from rigid links and passively compliant joints \cite{albu2004earlydef}, have become outdated. Some more recent definitions focus on quantitative measures of softness based on Young's modulus \cite{wang2015softmatter, rus2015design}, while others reject this to account for a broader scope of environmental interaction \cite{chen2017definition}. For the purpose of this review, we will define the term "soft robot" similarly to \cite{laschi2016science} and \cite{chen2017definition}, as a system which employs the inherent or structural compliance of its construction materials to actively interact with its environment.  Within this field, biological inspiration is often crucial to the systems which meet these criteria. Similarly to \cite{whitesides2018soft}, we refer to biologically inspired and biomimetic systems as systems which attempt to reproduce the function of a biological system, but not necessarily the underlying mechanisms. Interestingly, the evolution of the definition of soft robotics is coincident with the increasing interest in bioinspired systems, highlighting the interconnectedness of these topics.

As robots move outside of the structured environments of industrial settings and aim to assist in a growing range of task spaces \cite{acemoglu2020robots,laschi2016science}, the importance of rigid precision placed on conventional robotic systems is outweighed by the need for safe interaction with varying environments and the people within them. It is in large part this paradigm shift within the field which motivates interest in bioinspired soft robotics. By reevaluating system construction materials and developing new methods of actuation based on biological mechanisms, soft robots have been able to elevate capabilities in a wide spectrum of fields from surgical procedures \cite{diodato2018surgery} to ocean and space exploration \cite{aracri2021soft,peck2016fish}. In addition to system compliance, soft robots offer new and unique actuation mechanisms and in some cases nearly infinite degrees of freedom as well as resistance to damage, further differentiating themselves from rigid counterparts. 

At its inception, the field of soft robotics primarily concerned itself with the development of hybrid soft devices which facilitated safer physical interaction with humans for medical orthoses \cite{nickel1963development} or within the context of industry \cite{albu2008soft, coyle2018bio}. However, as the field and its definition evolved, nature and biological design helped guide the trajectory of research. For example artificial muscles
take inspiration from musculoskeletal systems \cite{greco2022muscles}. New methods of surface and underwater locomotion draw inspiration from crawling invertebrates and the fluid strokes of jellyfish and rays \cite{kim2013soft,coyle2018bio}. Muscular hydrostats such as elephant trunks and squid tentacles form a basis for new approaches to manipulation and grasping devices \cite{coyle2018bio, manti2016stiffening}. The examples of biological influence are numerous and broad reaching, leading to a wealth of research and impact for the field. 

It is the purpose of this review to provide an understanding of the state of the art for bioinspired soft robotic devices as well as a discussion of consistent modeling and control techniques which are used throughout the field. Also included will be discussion and remarks on the field's trajectory and the challenges faced by researchers.

\section{Actuation Approaches}

\begin{figure}
    \centering
    \includegraphics[width=\textwidth]{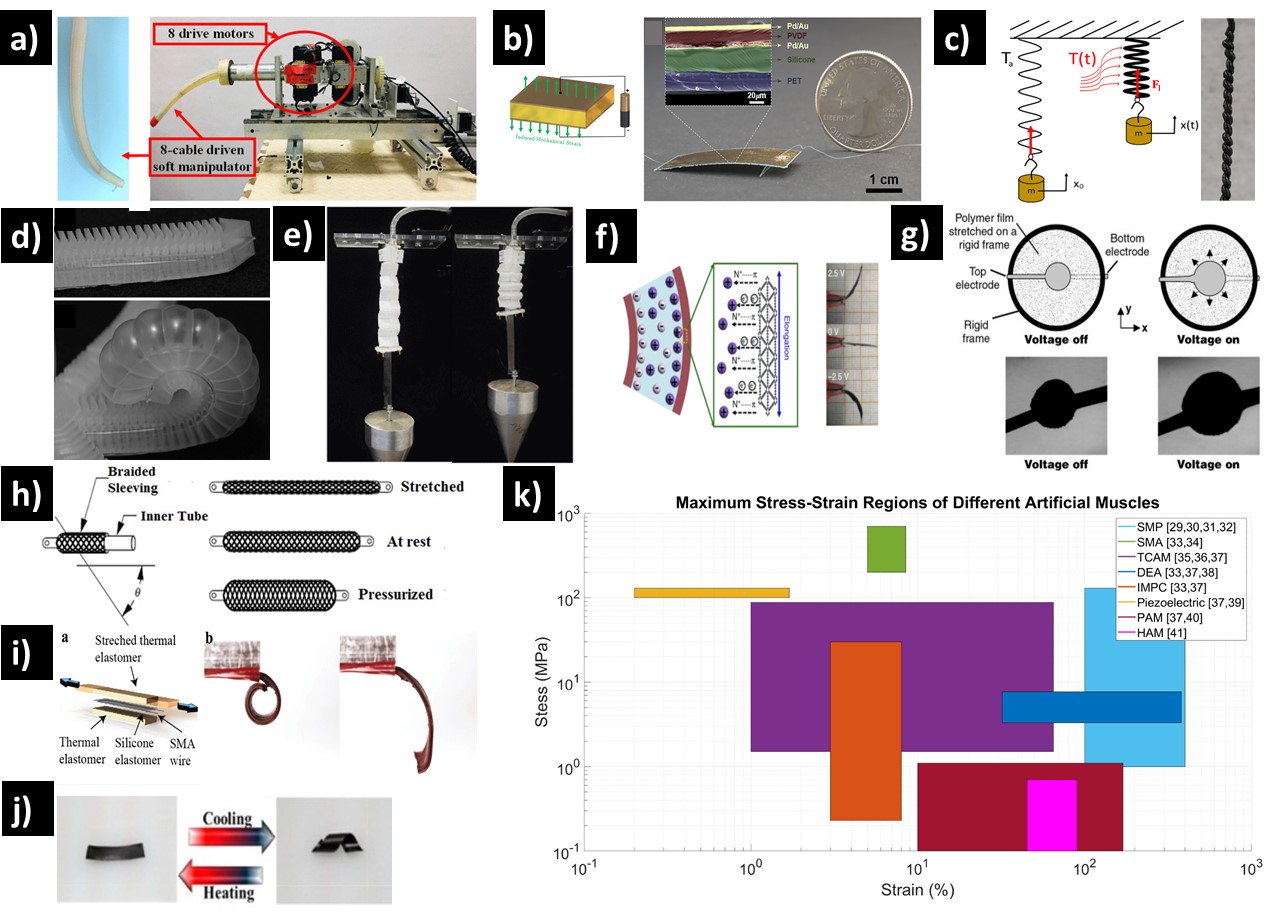}
    \caption{Visual summary of discussed actuation methodologies through examples and diagrams: a) cable driven device where a soft tentacle's shape is controlled by embedded cable tendons attached to electromagnetic motors (\textcopyright [2021] IEEE. Reprinted, with permission, from \cite{xu2021visual}); b) an illustration of the inverse piezoelectric effect (\textcopyright[2020] IOP Publishing, Ltd. Reprinted, with permission, from \cite{mohith2020recent}), and a insect scale crawling device made from a single piezoelectric artificial muscle (\textcopyright [2019] American Association for the Advancement of Science. Reprinted, with permission, from \cite{wu2019insect}); c) a carbon fiber silicone rubber TCAM with actuation achieved by anisotropic thermal expansion (\textcopyright [2019] IOP Publishing, Ltd. Reprinted, with permission, from \cite{giovinco2019dynamic}); d) an example of pnue-net actuation, inflating networks of void space within an elastomer to generate a desired shape (\textcopyright [2011] John Wiley \& Sons, Inc. Reprinted, with permission, from \cite{ilievski2011soft}); e) a 3D printed origami pnuematic actuator using a vacuum pump to lift a mass (\textcopyright [2019] IEEE. Reprinted, with permission, from \cite{zhang2019design}); f) IMPC working mechanism illustration showing ionic motion inside material causing elongation on one side of an actuator, and an example of a thin IMPC actuator working under applied voltage \cite{lu2018high}; g) a DEA actuator exploiting elctrostatic pressure from the DE material under applied voltage to stretch an elastomer membrane (\textcopyright [2015] Elsevier. Reprinted, with permission, from \cite{romasanta2015increasing}); h) braided PAM/HAM actuator diagram showing how pressurized fluid works with a braided sleeve to contract \cite{kalita2022review}; i) thermally actuated SMA wires used to control an elastomer's shape (\textcopyright [2019] John Wiley \& Sons, Inc. Reprinted, with permission, from \cite{huang2019highly}); j) SMP reversible thermal actuator moving between two memorized shapes based on thermal input (\textcopyright [2019] Elsevier. Reprinted, with permission, from \cite{zare2019thermally}); k) Artificial muscle stress-strain comparison based on table in \cite{greco2022muscles}. Reference values \cite{gall2004shape,small2010biomedical,miaudet2007shape,leng2011shape,madden2004artificial,foroughi2011torsional,chen2015hierarchically,mirvakili2018artificial,hartl2007aerospace,brochu2010advances,uchino2017advanced,daerden2002pneumatic,li2017fluid}.}
    \label{fig:actuators}
\end{figure}

Actuator design plays a crucial role in enabling soft robotic systems with a large number of degrees of freedom to interact with the environment and achieve a desired actuation pattern. This is especially true when attempting to mimic the intricacies of biological systems. In the past few decades, researchers have devoted a great deal of effort toward the development of light-weight and highly compliant actuation technologies. While some individual actuation mechanisms may themselves not appear bioinspired, they merit discussion as common components which enable biomimetic behavior on a system scale. Among these are fluidic elastomer actuators (FEA), hydraulic and pneumatic systems which pressurize inner networks of chambers within the soft material they are driving \cite{shintake2018soft,walker2020pneu}. Most common in this approach are pneu-nets \cite{wang2018programmable,gu2021analytical,natarajan2021bio,zhang2022herringbone,de2016constitutive,jiang2021modeling} (pictured in Figure \ref{fig:actuators}d), which are frequently used to enable crawling locomotion \cite{walker2020crawl,natarajan2021bio,shepherd2011multigait,tang2020development} as well as grasping and manipulation \cite{ilievski2011soft,huang2020variable,chen2018pneumatically,homberg2019robust}. Similar in concept, origami structures such as the one in Figure \ref{fig:actuators}e, which unfold and collapse in response to pneumatic input, are increasingly common in the field \cite{yu2020crawling,zhang2019design,paez2016design,son20224d}. Beyond such approaches, examples of cable-driven systems are also prevalent \cite{bern2019trajectory,wang2017cable,xu2021visual}, offering exact inputs to a system from electromagnetic motors and acting similarly to muscle-tendon systems observed in biology (Figure \ref{fig:actuators}a). In many cases, the aforementioned strategies offer high power and well characterized inputs, but come at the cost of heavy components and are often tethered by tubing or wires. Even so, these methods are able to shift weight away from the moving components of the robotic systems they power, providing benefits in more spatially confined tasks. 

Building upon these more classical actuation methods, artificial muscles, which take their inspiration from the high force-to-weight ratios and compact nature of biological muscle, have garnered increasing momentum in the field \cite{greco2022muscles,mirvakili2018artificial,zhang2019robotic,wang2021recent}. Defined as materials or devices that can reversibly actuate within a single component given specific external input \cite{greco2022muscles}, these actuators trend away from conventional energy inputs toward more specialized ones that exploit the specific material properties of a given artificial muscle. This class of actuator encompasses a broad swath of examples; however, observing the field through the lens of tethered and untethered actuators helps to narrow scope, and separate potential use cases \cite{wang2021recent}. 

Tethered artificial muscles suffer some of the same limitations of the more conventional methods already discussed (e.g. requiring power supplies or pneumatic compressors) but each offer unique benefits as well. Dielectric elastomer actuators (DEA) are one such actuator, capable of strains as high as 350\%, with stresses of several MPa \cite{naficy2016bio}. DEAs exploit Coulombic attraction between electrodes separated by a compressible membrane to actuate  \cite{kofod2001dielectric,hajiesmaili2021dielectric,romasanta2015increasing,o2008review} (as pictured in Figure \ref{fig:actuators}g), and can also be used in sensors and electrical power generators \cite{samatham2007active}. Similarly, ionic-polymer metal composites (IPMC) use electrodes to drive ion motion inside a polymer to cause volume change \cite{shahinpoor1998ionic} (shown in Figure \ref{fig:actuators}f). Both DEA and IMPC actuators belong to the larger family of electrically actuated elastomers (EAP), a collective term for active polymers which change shape under an applied voltage \cite{bar2004electroactive,bahramzadeh2014review,naficy2016bio,yang2020actuation,aabloo2020challenges}. EAPs are generally known for fast action and high power output at the cost of being tethered to their power supply. Pneumatic artificial muscles (PAM) \cite{nickel1963development,yang2016buckling,daerden2002pneumatic,kalita2022review} and hydraulic artificial muscles (HAM) \cite{tiwari2012hydraulic,li2017fluid} are other examples of tethered artificial muscles, trading power supplies for compressors and pumps. These muscles offer simplicity and high efficiency at low economic cost \cite{daerden2002pneumatic,tiwari2012hydraulic}, and an example of the braided variant of these muscles is shown in Figure \ref{fig:actuators}h. 

In the realm of untethered artificial muscles, common stimuli include electromagnetic fields, light, thermal radiation, and chemical reaction or swelling. Piezoelectric artificial muscle actuators can be tethered or untethered, producing mechanical work in an electric field, inverse to the piezoelectric effect (as shown in Figure \ref{fig:actuators}b). These muscles can produce precise, energy efficient motion at high speed, but require very high voltages and are often brittle \cite{zhang2019robotic,peng2013survey,shouji2023fast,mohith2020recent}. Other such examples are shape memory alloys (SMA) and shape memory polymers (SMP), which are materials that can recover a memorized shape from a number of different stimuli, most commonly thermal input \cite{huang2019highly,huang2018chasing,jin2016soft,mohd2017designing,jani2014review,ali2022novel,taniguchi2013flexible,xia2021review,zare2019thermally}. Examples of their use as actuators can be seen in Figure \ref{fig:actuators}i and \ref{fig:actuators}j. While capable of providing high power density and stress, these actuators are often limited by small strokes, significant hysteresis, and creep \cite{zhang2019robotic,zhang2018three}. Twisted and coiled artificial muscles (TCAM) \cite{haines2014artificial,madden2015twisted,lamuta2018theory,foroughi2011torsional} (Figure \ref{fig:actuators}c) and twisted and spiraled artificial muscles (TSAM) \cite{lamuta2019digital} are treated fibers that respectively compress or expand as a result of thermal expansion or chemical swelling. In the case of TCAMs and TSAMs, thermal influence can be provided by means of joule heating \cite{giovinco2019dynamic} which would tether them. These muscles are capable of strains up to 50\% and exhibit high force-to-weight ratios \cite{giovinco2019dynamic,haines2014artificial}, but are inhibited by limited speed and force outputs from individual muscles \cite{zhang2019robotic}. While the many advantages of untethered muscles are accompanied by drawbacks, the importance of their ability to enable unique actuation without limited reach enables many biomimetic systems to enter new environments. 

Figure \ref{fig:actuators}k provides a quick reference for the range of maximum stresses and strains of the discussed artificial muscles, highlighting further considerations for choosing system specific actuators, and the general trend of trade offs between muscle stroke and force output.  

\section{Soft Robot Functions}
As stated, the applicability of soft robots covers a large breadth, and the diversity of biological inspirations has allowed for multiple solutions to various problems. To provide a holistic view of the broad field in a concise review, the current research approaches to locomotion, grasping, and manipulation using soft robots will be the focus of this section.

\subsection{Locomotion}
Soft robotic approaches to locomotion can largely be separated by terrestrial, aerial, and aquatic devices \cite{calisti2017fundamentals,ng2021locomotion,pal2021exploiting}. 
In order to limit scope, hybrid soft robots made from primarily hard components which exploit small elastic elements will not be discussed. 

\begin{figure}
    \centering
    \includegraphics[width=\textwidth]{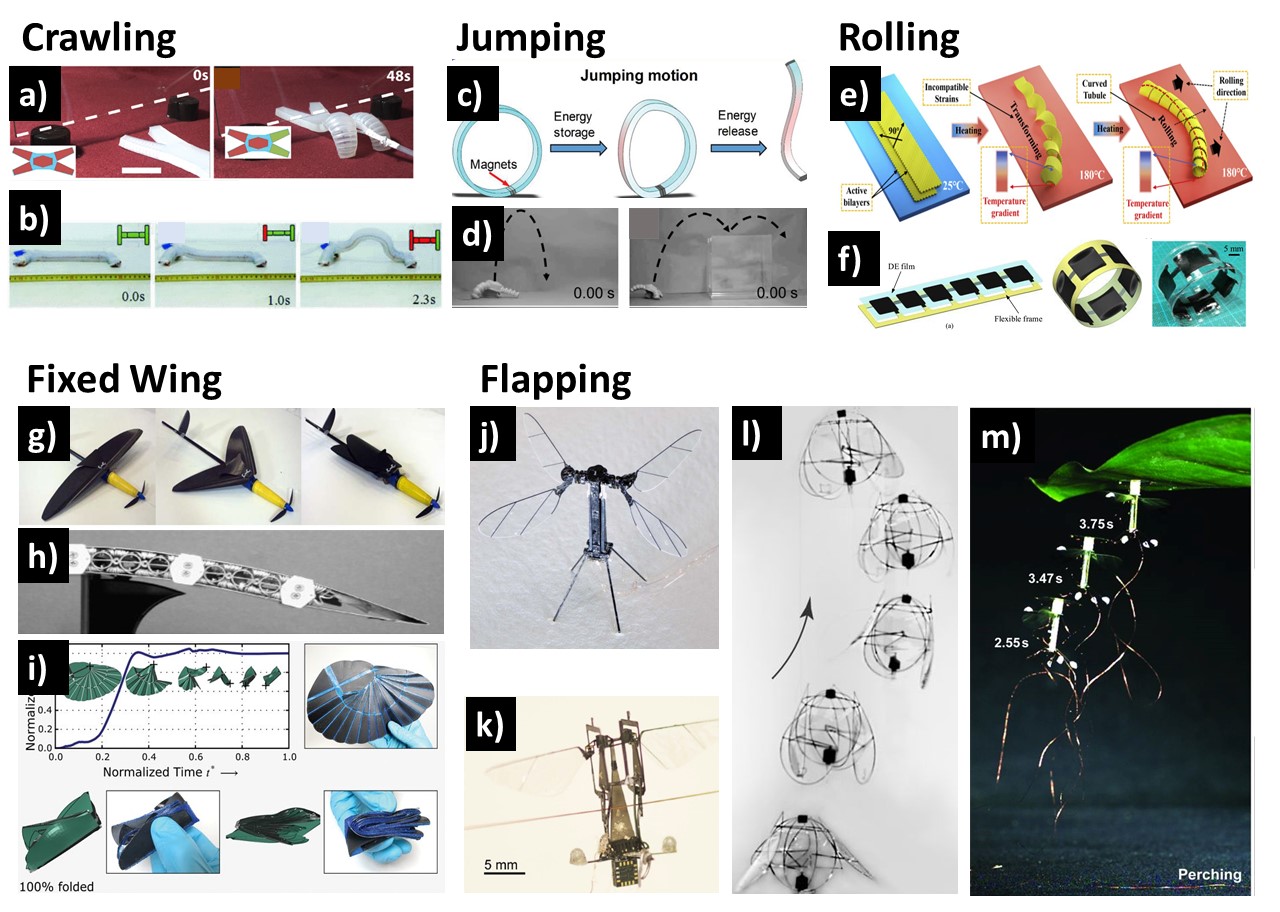}
    \caption{Terrestrial and aerial locomotion visual summary: a) a pneumatically actuated multigate crawling device demonstrating its ability to crawl under a short gap (\textcopyright [2011] PNAS. Reprinted, with permission, from \cite{shepherd2011multigait}); b) an illustration of the actuation sequence of two-anchor crawling for an inchworm-like device (\textcopyright [2018] IEEE. Reprinted, with permission, from \cite{Xie2018PISRob}); c) light responsive liquid crystal elastomer material jumping by storing and releasing energy (\textcopyright [2019] John Wiley and Sons, Inc. Reprinted, with permission, from \cite{ahn2019bioinspired}); d) unteathered pneumatic jumping device using combustion in its rear appendage to rapidly pressurize (\textcopyright [2014] IEEE. Reprinted, with permission, from \cite{tolley2014untethered}); e) a rolling device made with additive manufacturing which moves forward as it reshapes itself due to thermal input (\textcopyright [2021] Elsevier. Reprinted, with permission, from \cite{zhai20214d}); f) DEA powered rolling device, bending to shift its center of mass and move forward (\textcopyright [2018] IEEE. Reprinted, with permission, from \cite{li2018fast}); g) a fixed wing robot which folds its wings to perform plunge diving maneuvers (\textcopyright [2017] The Royal Society (U.K.). Reprinted, with permission, from \cite{siddall2017wind}); h) an example of wing morphology alteration where SMA are used to induce chord-wise bending of an airfoil  (\textcopyright [2010] Elsevier. Reprinted, with permission, from \cite{sofla2010shape}); i) insect inspired spring origami wing folding for fast deployment of wings and compact storage  (\textcopyright [2018] The American Association for the Advancement of Science. Reprinted, with permission, from \cite{faber2018bioinspired}); j) k) Robobee, a device using piezoelectric actuators to drive four wings ( j) \textcopyright [2019] Springer Nature. Reprinted, with permission, from  \cite{vaughan2019robobee}, k) \textcopyright [2014] IEEE. Reprinted, with permission, from \cite{helbling2014pitch}); l) jellyfish inspired flier which uses low frequency flapping to hover \cite{ristroph2014stable}; m) flapping wing insect robot which uses electrostatic adhesion for perching underneath surfaces (\textcopyright [2016] The American Association for the Advancement of Science. Reprinted, with permission, from \cite{graule2016perching}).}
    \label{fig:terra-aer}
\end{figure}

The first major mode of terrestrial locomotion is crawling. This can be separated further to: two-anchor crawling, similar to an inchworm \cite{Xie2018PISRob,ahn2019bioinspired,zhu2017architectures,schuldt2015template,rozen2021design} (Figure \ref{fig:terra-aer}b); peristaltic locomotion, whereby the motion is produced from radial contraction and axial elongation \cite{tang2020development,ze2022soft,zhu2017architectures,seok2010peristaltic,kandhari2021analysis}; serpentine crawling in which a wave of deformation passes through the robot similar to a snake \cite{onal2013autonomous,qi2022bioinspired,yang2019graphene}; multi-gate devices capable of multiple modes of crawling locomotion \cite{natarajan2021bio,sun2021soft,shepherd2011multigait,choi2022artificial} (Figure \ref{fig:terra-aer}a). A primary advantage of soft robotic terrestrial locomotion is the ability to traverse irregular surfaces by exploiting inherent compliance. This is especially apparent in the case of crawling robots, as these lightweight systems can maintain surface contact by means of deforming to fit their environment \cite{sugiyama2004crawling}. Another mode of ground traversal is jumping, where potential energy is built up in some elastic element (as seen in Figure \ref{fig:terra-aer}c) or reaction (as seen in Figure \ref{fig:terra-aer}d), and released to move the robot \cite{calisti2017fundamentals,tolley2014untethered,pal2021exploiting, ahn2019bioinspired}. This enables quick bursts of speed, which is desirable by contrast to the low speed of most soft locomotion, and allows robots to move over gaps and obstacles easily \cite{pal2021exploiting}.  Rolling soft robots are also common in this application, deforming at points of contact with the ground to control their trajectory \cite{li2018fast,zhai20214d,xiao2020liquid,fu2022humidity} (as exemplified in Figure \ref{fig:terra-aer}e and \ref{fig:terra-aer}f). This method offers increased and sustained speed relative to crawling or jumping, but limits application, and cannot overcome large obstacles or gaps in terrain \cite{ng2021locomotion}.

Aerial soft robots can generally be subdivided into fixed and flapping wing devices \cite{calisti2017fundamentals}, as shown in Figure \ref{fig:terra-aer}g-m. In the case of the former, compliant wing construction has been used to enable wing morphing, allowing for augmentation of flight behavior in several ways. Wing characteristics such as span, cord length, sweep \cite{blondeau2003wind,sofla2010shape,siddall2017wind}, flight envelope \cite{vasista2012realization}, and surface texture \cite{bartley2004development} have all been the targets of morphing devices to generate desired wing behavior. Gliding wings, taking inspiration from the folding membrane wings of bats and insects, have also been studied \cite{faber2018bioinspired,bleischwitz2015aspect}, allowing wing folding for storage and deployment. Flapping wing locomotion is the preferred method of many small scale robots as a result of the low Reynolds number regimes in which they operate making fixed wings impractical \cite{farrell2018review}. To this end, small scale elastic actuation systems have been developed alongside folding wings to enable this kind of flight \cite{whitney2010aeromechanics,zhao2010aerodynamic,tanaka2015flexible,chen2022analysis}. A particular benefit of this mode of locomotion is the ability to continuously accelerate and hover \cite{ng2021locomotion}.

\begin{figure}
    \centering
    \includegraphics[width=\textwidth]{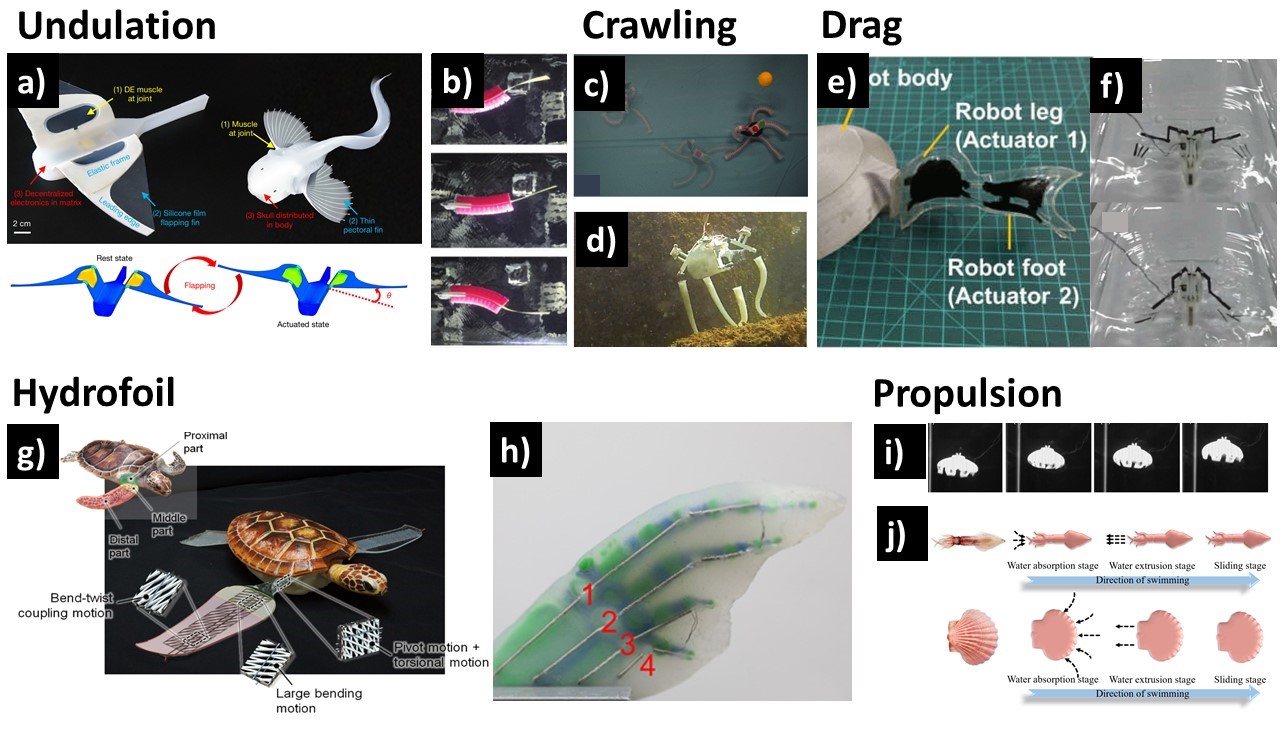}
    \caption{Aquatic locomotion visual summary: a) flapping fish robot using DE actuators to move fins to generate thrust and control depth (\textcopyright [2021] Mary Ann Liebert, Inc. Reprinted, with permission, from \cite{li2021self}); b)  undulating fish tail actuated by pneu-nets (\textcopyright [2017] Mary Ann Liebert, Inc. Reprinted, with permission, from \cite{jusufi2017undulatory}); c) starfish crawling device with SMA actuated tentacles which drag the robot (\textcopyright [2020] IEEE. Reprinted, with permission, from \cite{patterson2020untethered}); d)  multi-modal octopus demonstrating bethic crawling locomotion with electric motor driving the tentacle legs (\textcopyright [2017] IOP Publishing, Ltd. Reprinted, with permission, from \cite{giorgio2017hybrid}); e) DEA leg of a biomimetic frog swimming robot exploiting webbed feet to increase drag (\textcopyright [2017] IEEE. Reprinted, with permission, from \cite{tang2017frog}); f) insect swimming robot using DC motor actuated oars to induce drag \cite{kwak2016design}  g) 3D printed SMA in biomimetic turtle fins to create different actuation based on scaffold structures (\textcopyright [2016] IOP Publishing, Ltd. Reprinted, with permission, from \cite{song2016turtle}); h) a dorsal fin actuated by coiled polymer artificial muscles to influence drag, lift, and laminar-to-turbulent transitions of the hydrofoil \cite{hunt2019soft}; i) jellyfish robot demonstrating  propulsion lift sequence \cite{villanueva2011biomimetic}; j) a diagram showing clam and squid propulsion processes to be recreated by robotics (\textcopyright [2020] IOP Publishing, Ltd. Reprinted, with permission, from \cite{wang2020development}).}
    \label{fig:aqua}
\end{figure}

The aquatic locomotion of soft robots has a number of applications and methodologies depending on the system's scale. At the micro to millimeter length, robots can be used in minimally invasive medical procedures to deliver drugs in hard to reach places \cite{ceylan20193d} or for small scale manufacturing in fluids \cite{shields2010biomimetic,sugioka2020artificial}. In many of these cases, the biology of small organisms such as cells \cite{shields2010biomimetic,sugioka2020artificial} and jelly fish \cite{ren2019multi} are the inspiration for actuation, using artificial cilia or propulsion to move. On a larger scale, the bioinspired characteristics of aquatic soft robotic locomotion is largely made up by a combination of lift from hydrofoils, undulation, drag, and jet propulsion \cite{calisti2017fundamentals}, and some examples can be seen in Figure \ref{fig:aqua}. In the case of hydrofoils, biological inspiration from the fins and flippers of different sea creatures, such as rays \cite{chen2012bio,park2016phototactic} and turtles \cite{song2016turtle}, lead to an array of designs. These structures can change characteristics similarly to the wings used in cases of aerial locomotion, providing desired hydrodynamic force and forward thrust to the system. The undulatory swimming motion of fish has been well studied as a means to provide propulsion to aquatic robots \cite{jusufi2017undulatory,wolf2020fish}, and can be used in combination with hydrofoils, as seen in the case of animals such as rays \cite{park2016phototactic}. In the context of aquatic locomotion, drag is employed by animals such as frogs and insects, which use their limbs like oars to move \cite{tang2017frog,kwak2016design}. Similar to aerial flapping, locomotion by exploiting drag allows desirable attributes for precise control and hovering in liquid environments. Inspiration for jet propulsion in soft aquatic robots comes from a variety of biological systems such as jellyfish, cephalopods, and scallops \cite{villanueva2011biomimetic,renda2015structural,wang2020development}, lending benefits of large acceleration and manoeuvrability to soft robots which employ this methodology. Underwater surface crawling is also well represented in aquatic locomotion, often drawing inspiration from octopuses and starfish \cite{patterson2020untethered,hermes2021bioinspired,giorgio2017hybrid} as shown in the examples of Figure \ref{fig:aqua}c and \ref{fig:aqua}d, furthering exploratory capabilities.  

\subsection{Grasping}

\begin{figure}
    \centering
    \includegraphics[width=\textwidth]{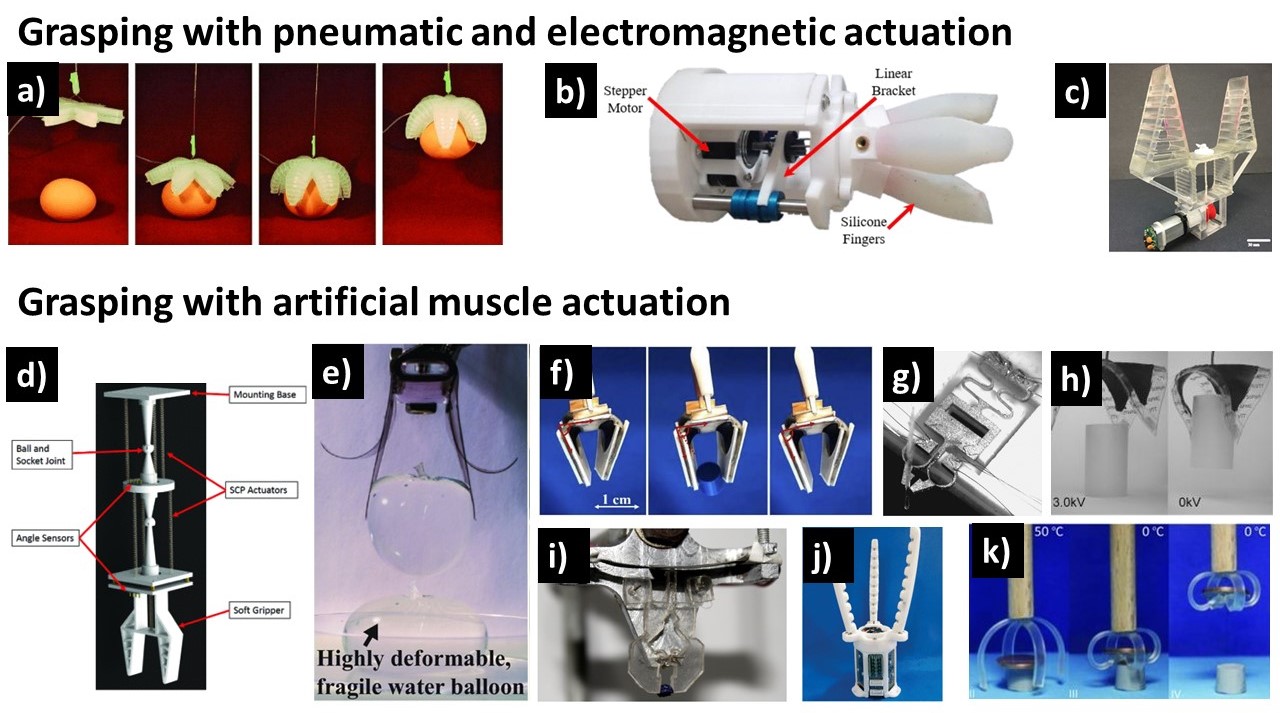}
    \caption{Visual summary for grasping soft robots separating conventional and artificial muscle actuation: a) the starfish gripper, one of the first examples of pneu-net actuation implemented to enable grasping of an egg (\textcopyright [2011] John Wiley \& Sons, Inc. Reprinted, with permission, from \cite{ilievski2011soft}); b) tendon driven soft fingers actuated by a stepper motor allowing delicate berry picking to be performed (\textcopyright [2022] IEEE. Reprinted, with permission, from \cite{gunderman2022tendon}); c) compliant structure powered by electromagnetic motor for delicate interaction \cite{crooks2016fin}; d) SCP driven gripper and arm with similar compliant structure to c at the end effector (\textcopyright [2020] IEEE, Inc. Reprinted, with permission, from \cite{yang2020compact}); e) DEA driven gripper using electrostatic adhesion to enhance grasping capabilities (\textcopyright [2016] John Wiley \& Sons, Inc. Reprinted, with permission, from \cite{shintake2016versatile}); f) multilayer DEA driven gripper working to overcome gasping force limitations of single layer devices \cite{thongking2021soft}; g) micro scale gripper created from SMA (\textcopyright [2002] Elsevier. Reprinted, with permission, from \cite{kohl2002sma}); h) DEA driven gripper designed using energy minimization techniques (\textcopyright [2007] AIP Publishing. Reprinted, with permission, from \cite{kofod2007energy}); i) millimeter scale IMPC actuated PDMS gripper (\textcopyright [2015] Springer Nature. Reprinted, with permission, from \cite{bhattacharya2015simultaneous}); j) SCP driven gripper using actuator thermal model for self-sensing (\textcopyright [2021] IEEE, Inc. Reprinted, with permission, from \cite{wang2021lightweight}); k) thermally actuated SMP driven gripper lifting a quarter (\textcopyright [2013] John Wiley \& Sons, Inc. Reprinted, with permission, from \cite{behl2013reversible}).}
    \label{fig:grasping}
\end{figure}

Grasping is fundamental to robotic environmental interactions \cite{birglen2008grasping}, and softness in grasping implements allows for desirable outcomes when handling delicate or irregularly shaped objects. The broadening spectrum of potential interactions afforded by soft grasping systems has sparked significant investigation into the subject, and many approaches have been taken in the design of new grasping systems \cite{shintake2018soft}. 
One such approach was utilized by the starfish gripper, which employs pneu-nets, and demonstrated soft handling of delicate food items by conforming to their shape due to inherent system compliance \cite{ilievski2011soft,whitesides2018soft}. The success of this design, the simplicity of its manufacturing, along with its clear functional benefits, have led to further investigation into FEA driven grippers \cite{huang2020variable,chen2018pneumatically,homberg2019robust}. Alongside this approach, passive structures driven by external electromagnetic motors are well represented, commonly seen as grippers with tendon actuation or contact-driven deformation. Tendon driven grasping systems \cite{birglen2008grasping,gunderman2022tendon,jeong2021reliability,mutlu20163d}, inspired by human fingers, are underactuated structures which close through application of tension to a tendon. While electric motors are commonly used to supply that tension, examples of artificial muscles in these systems also exist \cite{ganguly2012control,phan2020hfam,wu2017compact}. Similarly, passively compliant structures at a gripper's points of contact can be actuated in fully soft \cite{crooks2016fin}, and combined soft/rigid systems \cite{nishimura2017variable,guo2017design}. In both cases, these passive structures offer more delicate capabilities to systems which follow many conventions of rigid robotics by enabling similar shape conformance as that seen in the FEA gripper case. These approaches are visualized in Figure \ref{fig:grasping}a-c.
 
In addition to those grasping systems employing electric motors and pnuematic devices for actuation, artificial muscles have been investigated for use in grasping devices as well. Often, this enables significant simplifications to a grasping device's design, combining the grasping elements and actuation into one component. For example, 
DEA grasping devices have been demonstrated as single-component grippers \cite{kofod2007energy,araromi2014rollable}, typically using pretensioned DEA on a flat flexible frame to create a desired actuation. Current research on these devices focuses on overcoming their low generated grasping forces through multilayer stacking \cite{thongking2021soft}, electrostatic chucking \cite{imamura2017variable}, and other methods \cite{shintake2016versatile,guo2018soft, shintake2018soft}. Similar system simplifications have been demonstrated with SMP \cite{behl2013reversible,bharti2016formation} and IMPC \cite{bhattacharya2015simultaneous,hamburg2016soft,kim2014bio} actuation however material limitations such as low recovery stress in the case of the former, and slow response in the latter have limited recent interest. In the case of SMAs, their relatively high active stress and scalability have enabled several small scale grasping devices \cite{zhong2006development,zhou2022modeling,kohl2002sma,kyung2008design} which is difficult to achieve through other methods. In addition to the single component gripper cases, super-coiled polymers (SCP), similar artificial muscles to the discussed TCAMs, have garnered attention as gripper actuators, as they help to overcome some of the issues of small stroke and loading limits of EAP grippers \cite{wang2021lightweight,yang2020compact}; however, systems employing them are similarly complicated to those tendon actuation systems. Examples of the described systems can be seen in Figure \ref{fig:grasping}d-k.

\subsection{Manipulation}
\label{sect:manip}

\begin{figure}
    \centering
    \includegraphics[width=\textwidth]{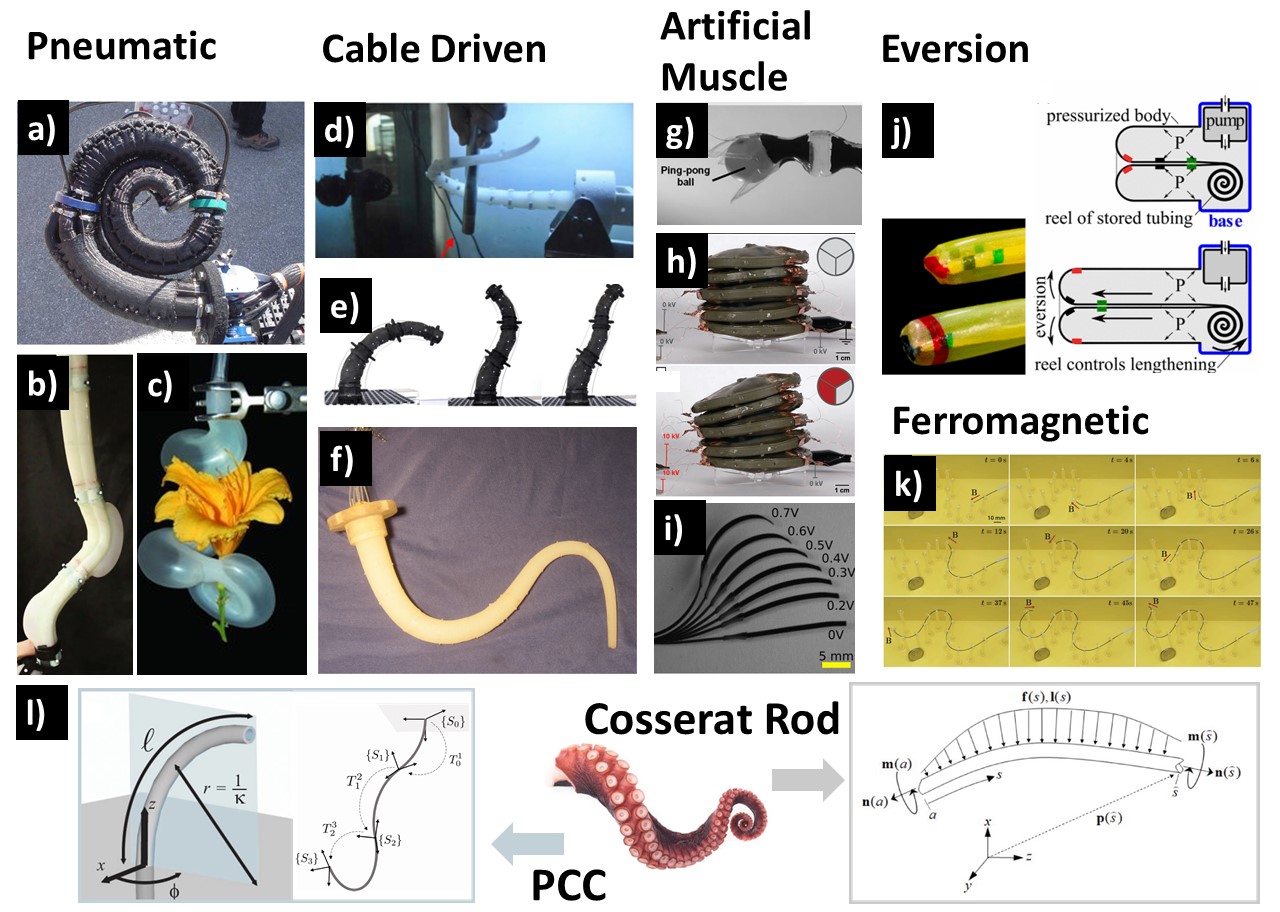}
    \caption{Visual summary for soft manipulation devices separated by actuation method: a) Braided pneumatic artificial muscle manipulator with rigid section division (\textcopyright [2006] IEEE. Reprinted, with permission, from \cite{mcmahan2006field}); b) FEA segmented manipulator with gripper end effector (\textcopyright [2016] SAGE Publications. Reprinted, with permission, from \cite{marchese2016design}); c) FEA manipulator with delegate grasping capabilities (\textcopyright [2013] John Wiley \& Sons, Inc. Reprinted, with permission, from \cite{martinez2013robotic}); d) cable actuation tentacle with SMA placed in cross its cross section for elongation (\textcopyright [2012] IOP Publishing. Reprinted, with permission, from \cite{mazzolai2012soft}); e) cable driven manipulator made from soft components with rigid divisions between sections \cite{wockenfuss2022design}; f) cable driven elastomer manipulator with embedded tendons (\textcopyright [2014] IEEE. Reprinted, with permission, from \cite{renda2014dynamic}); g) three dimensional segmented manipulator actuated by DEA with an attached gripper (\textcopyright [2020] Mary Ann Liebert, Inc. Reprinted, with permission, from \cite{xing2020super}); h) HASEL actuated manipulator segment exploiting layer expansions to generate larger scale shape control (\textcopyright [2020] John Wiley \& Sons, Inc. Reprinted, with permission, from \cite{o2020rapid}); i) planar manipulator made from thin DEA actuated segments (\textcopyright [2018] IEEE. Reprinted, with permission, from \cite{benouhiba2018multisegment}); j) Eversion actuated extendable manipulator which elongates for exploration of confined environments (\textcopyright [2018] IEEE. Reprinted, with permission, from \cite{greer2018obstacle}); k) submillimeter scale ferromagnetic manipulator capable of precise actuation in response to magnetic fields (\textcopyright [2019] The American Association for the Advancement of Science. Reprinted, with permission, from \cite{kim2019ferromagnetic}). l) Schematic representation showing the main differences between PCC and Cosserat rod based approaches. PCC separates the finite dimensional rod into discrete segments, each with given curvature characteristics, and the sum of these segments approximates the shape of the tentacle. The cosserat rod evaluates the tentacle continuously, with changing curvature and shear across the length of the rod. (Images: PCC \textcopyright [2010] SAGE Publications. Reprinted, with permission, from \cite{webster2010design}, \textcopyright [2019] IEEE. Reprinted, with permission, from \cite{katzschmann2019dynamic}; Cosserat \cite{janabi2021cosserat})}
    \label{fig:manip}
\end{figure}

Within the field of robotics, exactly defining manipulation can be somewhat challenging. Broadly, it refers to a system's ability to control elements of its environment through selective contact \cite{mason2018toward}. This is a step beyond grasping, referring to a robot's capabilities to move, reorient, or otherwise use objects they interact with \cite{mason2018toward,billard2019trends}. With the etymological roots of the word stemming from the Latin for hand, it makes sense that many of the existing soft robotic manipulators are inspired by the human hand \cite{puhlmann2022rbo,abondance2020dexterous,zhou2018bcl,lu2019soft,gupta2016learning}. These kinds of robots can be used for grasping, and towards controlling spacial orientation in limited domains; however, they are predominantly end-effectors which rely on other, often rigid, arms to reach their targets. These arms, with attached instruments for environmental interactions, are often what is being referred to when robots are called manipulators \cite{dou2021soft}. 

The field of robotic manipulation is well defined for rigid robotics \cite{lee2020critical,alandoli2020critical}, but several challenges still remain for soft systems. In this field, many soft devices build on the foundational work of continuum manipulators \cite{walker2013continuous}, a class of robots taking inspiration from octopus tentacle and elephant trunk physiology, which deform about a central cord to grant themselves nearly infinite degrees of freedom. Still, a large amount of these manipulators could be considered to have near rigid cross sections \cite{mcmahan2006field,trivedi2007geometrically,alambeigi2016continuum,gravagne2002uniform}, taking away from the intended goal of safely entering and maneuvering in cluttered environments \cite{dou2021soft,singh2014continuum}. This is often necessary to enable adequate precision in the presence of deflection resulting from gravity and other external loading \cite{trivedi2008soft}. Nevertheless, the promising potential of this class of robot drives research forward. As has been shown to be the case in soft robotics, a range of approaches are being taken to this end. 

Pnuematic actuation in the form of McKibben artificial muscles \cite{mcmahan2006field,trivedi2007geometrically}, pnue-nets \cite{katzschmann2015autonomous}, and other kinds of FEA \cite{marchese2014whole,martinez2013robotic,marchese2016design,cianchetti2014soft,fu2020interfacing} have shown varying degrees of success, with some examples confined to planar motion \cite{katzschmann2015autonomous,marchese2016design} and others offering more capabilities at the cost of softness \cite{mcmahan2006field,trivedi2007geometrically}. Tendon driven manipulators also exist \cite{renda2014dynamic,camarillo2008mechanics,feng2021learning,wockenfuss2022design,calisti2011octopus}, pulling on discrete connection points along the length of the continuum arm to cause bending, emulating the longitudinal muscles of tentacles. Some systems have also exploited both tendon and pneumatic actuation to control stiffness along the arms length in addition to position \cite{lee2016development,shiva2016tendon}. In these cases, the number of controllable degrees of freedom are limited by the number of discrete connection points of cables, or the inner geometry of inflatable chambers; however, they still clearly surpass the capabilities of conventional rigid link robots in this regard. More recently, alternatives to these well researched methods of actuation and construction have started to appear. Artificial muscle actuated systems using EAPs \cite{xing2020super,benouhiba2018multisegment}, hydraulically amplified self-healing electrostatic (HASEL) actuators \cite{o2020rapid,mitchell2019easy}, SMAs \cite{mazzolai2012soft} and SCPs \cite{yang2019novel} exist, but research in the area is quite young. Along side these example, unique actuation mechanisms such as pressure driven eversion (inspired by plants) \cite{hawkes2017soft,greer2018obstacle} and ferromagnetic composite ink \cite{kim2019ferromagnetic,wang2021evolutionary} have also emerged, enabling precise exploratory capabilities in delicate and small environments. A variety of examples of these soft continuum manipulators are given in Figure \ref{fig:manip}. Given the trajectory of research, it seems that it is in smaller scales that many of the advantages of soft manipulation systems begin to outweigh those of rigid counterparts, as unique actuation mechanisms enable robots to enter new task spaces.

\section{Modeling and Control}
\label{sect:model}

Given the similarities in the design of many bioinspired soft robots, stemming from attempts to replicate continuously deformable creatures such as snakes, or appendages such as octopus tentacles and elephant trunks, there are significant overlaps in modeling and control practices between many of the systems encompassed in the field. The importance of these topics to the implementation of robotic systems lends to a need for understanding common techniques used in this subsection of the field. Finite element methods (FEM) and finite dimensional models, such as piecewise constant curvature (PCC) or Cosserat rod models (visualized in Figure \ref{fig:manip}i), are among the most common approaches to modeling in the field, and they are often used to aid in the development of control methodologies. 

The modeling of elastic and hyperelastic materials using FEM is a field with a long history \cite{charlton1994review,weiss1996finite}. These methods have continued to be used for soft robotics, predominantly to aid in the design of structures to obtain desired behaviors, and to ensure performance and controllability \cite{zheng2019controllability,runge2017framework,smith2022stretching,naughton2021elastica}. Recent research in this area puts large emphasis on computational efficiency \cite{smith2022stretching,pozzi2018efficient}, which has even enabled realtime observers and control strategies using reduced order finite element models \cite{duriez2013control,tonkens2021soft,largilliere2015real,zhang2016kinematic}. This is desirable, especially in the case of robots with atypical designs \cite{duriez2013control,tonkens2021soft}; however, the high dimensionality of FEM models makes controller formulation difficult, and given the unique geometries of individual designs, general practices are hard to establish.

Alongside these developments, finite dimensional models have been another well documented approach to simulate flexible systems with high aspect ratios \cite{della2021model}, a category that encompasses various biomimetic soft robots. These models each use a similar underlying concept, simplifying a slender system into a finite dimensional rod which can be characterized by the position and orientation of any point along its arc length. The simplified nature of these models, which ignores intricate geometry that FEM can account for, makes them more suited for controller development than design. 

Among these methods, PCC models \cite{webster2010design,zhong2021bending,della2020improved} are some of the easiest to implement. A PCC model discretizes a rod along the arc length, and works under a few strong assumptions: strain is constant along the arc length; the curvature in any given discretized section is equivalently distributed; sections connect such that they are continuously differentiable \cite{della2021model,webster2010design}. These assumptions lead to key advantages for this approach. First, the resulting system of equations is expressed as an ordinary differential equation (ODE), which is generally easier to work with than the partial differential equations (PDE) common in modeling continuum systems. Second, given a further assumption that local curvature occurs in one plane, PCC models can be viewed as an extension of serial manipulators, where curvature sections are analogous to revolute joints and links \cite{della2021model}. This modeling technique has been used for the formulation of several kinds of feedback controllers. Example of these primarily include variations on PD controllers to regulate tip position or curvature angle \cite{katzschmann2019dynamic,falkenhahn2015model,della2019control,della2020improved,wang2021lightweight}, but examples of shape planning \cite{kolpashchikov2022fabrikx} and optimal control \cite{stella2022experimental,ding2022adaptive} have also recently emerged. It is notable that in the case of PD controllers, gain is often limited, and the integral term is neglected to maintain the inherent system softness \cite{katzschmann2019dynamic}. 

Alternative to PCC, the Cosserat rod model, which extends from the classical Kirchhoff rod model \cite{coleman1993dynamics}, is a set of PDEs capable of capturing all modes of deformation along a rod \cite{till2019real,janabi2021cosserat,spillmann2007corde,lang2011multi,renda2018unified}. This model's capabilities are increasingly desirable as more capable continuum systems emerge. Many control systems based on Cosserat theory design input force profiles necessary for obtaining desired system shape using infinite dimensional state feedback control \cite{Zheng2022PDE}, sensory feedback \cite{wang2022sensory}, slide mode control \cite{alqumsan2019robust}, model-based machine learning \cite{thuruthel2018model,pique2022controlling},  and principles of optimal control \cite{till2017Elastic,boyer2022statics,chang2022energy,chang2021controlling,wang2021optimal}, to name a few. Many of these controllers are tested only in simulation, as construction of systems capable of creating desired force profiles is still a topic of active research (see Section \ref{sect:manip}).

\section{Discussion}

Despite its youth, the field of biomimetic soft robotics has demonstrated significant staying power in the larger scheme of robotics \cite{bao2018bibliometric}. As evidenced by this review, the development of bioinspired soft systems has enabled weight reduction, system miniaturization, and unique actuation which vastly exceed the scope of conventional systems, and research trends suggest more to come. Further development fabrication process using additive manufacturing is set to ease soft robotic production and facilitate more complex designs \cite{stano2021additive}, and self healing materials promise more longevity to soft systems through resilience to damage and degradation \cite{roels2022processing,terryn2021review}. Even so, several challenges remain in many parts of the field, due in some part to the difficulties associated with replicating efficient biological actuation. Continued materials research and the further investigation of artificial muscles is important to close the gap between soft technologies and biology. Additionally, the youth of the field can be seen in the need for improved design tools. While FEM is being developed with pneumatic systems in mind \cite{smith2022stretching}, optimizing the design of soft robots is still a process in its very early stages, partially as a result of still developing models for the wide variety of newly available materials \cite{chen2020design}. Broadly applicable simulation efforts could aid in the design process over time, streamlining existing processes. Tangentially, continued improvement in reducing computational cost in modeling and simulation is important to efficient controller design and implementation on soft systems \cite{mengaldo2022concise}. Numerical methods for efficient solutions to models, such as the Cosserat rod model, could reduce system computational needs enabling weight reduction in on board computers. Additionally, further investigation into unteathered designs for soft continuum robots, which may be able to create desired force profiles along their length, is necessary in order to fully benefit from the implementation of higher end controllers. 

\section{Conclusion}

This review has provided an overview of the state of the art in bionspired soft robots with the hope of stimulating innovative and interdisciplinary research advances in this field.

\section{Acknowledgements}
This work was supported by DARPA (Young Faculty Award grant no. W911NF2110344-0011679424), ONR (grant no. N00014-20-1-2224, grant no. N00014-22-1-2021, and Young Investigator Program no. N00014-23-1-2116), and Iowa NASA EPSCoR Competitive Research Network (subaward no. 025372A from Iowa state University)

\section{Conflicts of Interest}
The authors declare that they have no conflict of interest.
\section{Human and Animal Rights and Informed Consent}
This article does not contain any studies with human or animal subjects performed by any of the authors.

\bibliography{refs}

\end{document}